\def\year{2020}\relax
\documentclass[letterpaper]{article} 
\usepackage{aaai20}  
\usepackage{times}  
\usepackage{helvet} 
\usepackage{courier}  
\usepackage[hyphens]{url}  
\usepackage{graphicx} 
\usepackage{graphicx}  
\newcommand\ourmodel{RankTxNet}
\usepackage{amsmath}
\usepackage{amssymb}
\usepackage{booktabs}
\usepackage{multirow}
\usepackage[caption=false]{subfig}

\title{Deep Attentive Ranking Networks for Learning to Order Sentences}
\author{Pawan Kumar\thanks{Equal contributions from both authors.}, Dhanajit Brahma$^{*}$, Harish Karnick, Piyush Rai
\\ Department of Computer Science and Engineering, IIT Kanpur, India\\ \{kpawan,dhanajit,hk,piyush\}@cse.iitk.ac.in}

\begin{document}

\maketitle

\begin{abstract}
We present an attention-based ranking framework for learning to order sentences given a paragraph. Our framework is built on a bidirectional sentence encoder and a self-attention based transformer network to obtain an input order invariant representation of paragraphs. Moreover, it allows seamless training using a variety of ranking based loss functions, such as pointwise, pairwise, and listwise ranking. We apply our framework on two tasks: Sentence Ordering and Order Discrimination. Our framework outperforms various state-of-the-art methods on these tasks on a variety of evaluation metrics. We also show that it achieves better results when using pairwise and listwise ranking losses, rather than the pointwise ranking loss, which suggests that incorporating relative positions of two or more sentences in the loss function contributes to better learning.
\end{abstract}

\section{Introduction}
Coherence is a fundamental aspect of natural language discourse and text.
In a coherent discourse, normally, sentences should respect the chronological order of events.
Correct logical ordering of parts of the discourse can facilitate understanding.
Ordering of sentences in a discourse determines local coherence, so it is an essential aspect of natural language processing.
The sentence ordering task tries to organize randomly shuffled sentences of a paragraph into a coherent text. Table~\ref{example-table} shows an Example of this task.
\cite{barzilay2008modeling} proposed the sentence ordering problem on the Accidents and Earthquakes datasets.

 Two of the most successful recent sentence representation models, Quick Thought~\cite{logeswaran2018efficient} and BERT (Bidirectional Encoder Representations from Transformers)~\cite{bert} use next sentence classification for learning the sentence representation.
 Next sentence classification is a special case of the sentence ordering task, which shows how vital the sentence ordering task is.
 Recently sentence ordering has been used in many applications like concept-to-text~\cite{konstas2012concept}, question answering~\cite{yu2018qanet,verberne2011retrieval}, multi-document summarization~\cite{barzilay2002inferring,nallapati2017summarunner}.

\begin{table}[t]
\centering
\resizebox{0.47\textwidth}{!}{%
\begin{tabular}{lc|ccc} \toprule
\multicolumn{1}{l}{} & \multicolumn{1}{l}{Unordered Paragraph} & \multicolumn{1}{|l}{} & \multicolumn{1}{l}{Ordered Paragraph} \\
\cmidrule{1-4}
3 & Then a nice thing happened. & 1 & Mario had lost his watch. \\ 
2 & He sent a mail to Lost \& Found. & 2 & He sent a mail to Lost \& Found. \\ 
4 & Somebody found his watch. & 3 & Then a nice thing happened. \\ 
1 & Mario had lost his watch. & 4 & Somebody found his watch. \\ 
\bottomrule
\end{tabular}
}
\caption{Example of unordered sentences in a paragraph (left) and ordered sentences in the same paragraph (right).}
\label{example-table}
\end{table}

Existing state-of-the-art for sentence ordering methods, such as \cite{wang2019hierarchical}, rely on sentence encoding with Transformer and Long Short-Term Memory \cite{Hochreiter1997LSTM} (LSTM) with pre-trained GloVe \cite{pennington2014glove} word vectors. These methods usually require sentence-by-sentence decoding to producing the reordered sentences. 

We propose a novel architecture for sentence ordering and reframe the problem in a \emph{ranking} framework. Our framework has several appealing properties.
Firstly, while most existing works~\cite{li2014model,gong2016end,chen2016neural,logeswaran2018sentence,cui-etal-2018-deep,wang2019hierarchical} use pre-trained word representations, we leverage Transformer based BERT sentence representations, allowing our model to use improved sentence encoding. The sentence encoder uses only Transformer (no LSTMs used).
Following previous works we use order invariant Transformer based paragraph encoder for paragraph encoding.
Secondly, while many recent works use a Pointer Network based decoder for decoding the sentence order from the encoded sentences, one sentence at a time, we propose a simple and efficient feed-forward neural network decoder.
It computes a relevance score for each sentence, in parallel.
These scores can be simply sorted to predict the correct sentence ordering (without expensive beam search).
Thirdly, predicting scores for every sentence in this fashion allows us to reframe the sentence ordering problem as a ranking problem.  
Our sentence ordering model can, thus, leverage extensive prior work on the \emph{Learning to Rank} framework~\cite{burges2005learning}.

We conduct an extensive evaluation of our model on six benchmark datasets for the sentence ordering task. We evaluate our model on two standard metrics: (1) Kendall's tau ($\tau$), (2) Perfect Match Ratio (PMR). We surpass the state-of-the-art models in terms of $\tau$ on all benchmark datasets. We also give our model's performance in terms of PMR on all the datasets.
Our model excels in making accurate first and last sentence predictions, achieving better performance than previous state-of-the-art approaches. We also provide visualizations of sentence representation and sentence level attention. On the order discrimination task, we show improvements over current state-of-the-art on Accidents dataset and give competitive results on Earthquakes dataset.

\section{\ourmodel{}: Deep Attentive Ranking Networks for Learning to Order Sentences}
This section starts with the sentence ordering problem set-up.
Then we describe the proposed model \ourmodel{}, which as building blocks uses BERT for sentence encoding and a Transformer for paragraph encoding.
We also give the details of training the model for various ranking loss functions.
\subsection{Problem Set-up}
The problem of sentence ordering deals with finding the correct order of sentences given a randomly ordered paragraph.
In other words, the aim is to find the most coherent permutation of sentences among all possible orders in a paragraph.
Given a paragraph $\mathcal{\mathbf{p}}=[s_{o_1},s_{o_2},\cdots,s_{o_m}]$ with $m$ sentences and order $\mathbf{o}=[o_1,o_2,\cdots,o_m]$, paragraph  $\mathbf{p^*}=[s_{o_1^*},s_{o_2^*},\cdots,s_{o_m^*}]$ is correctly ordered if $\mathbf{o^*}=[o_1^*,o_2^*,\cdots,o_m^*]$ is the order of the most coherent permutation of sentences.
For example in Table~\ref{example-table}, the unordered paragraph is $\mathbf{p}=[s_{o_1},s_{o_2},s_{o_3},s_{o_4}]=[s_3,s_2,s_4,s_1]$ and the correct order is $\mathbf{o^*}=[1,2,3,4]$.
\subsection{Model Overview and Intuition}
\begin{figure*}
    \centering
    \subfloat{
        \includegraphics[width=0.65\textwidth]{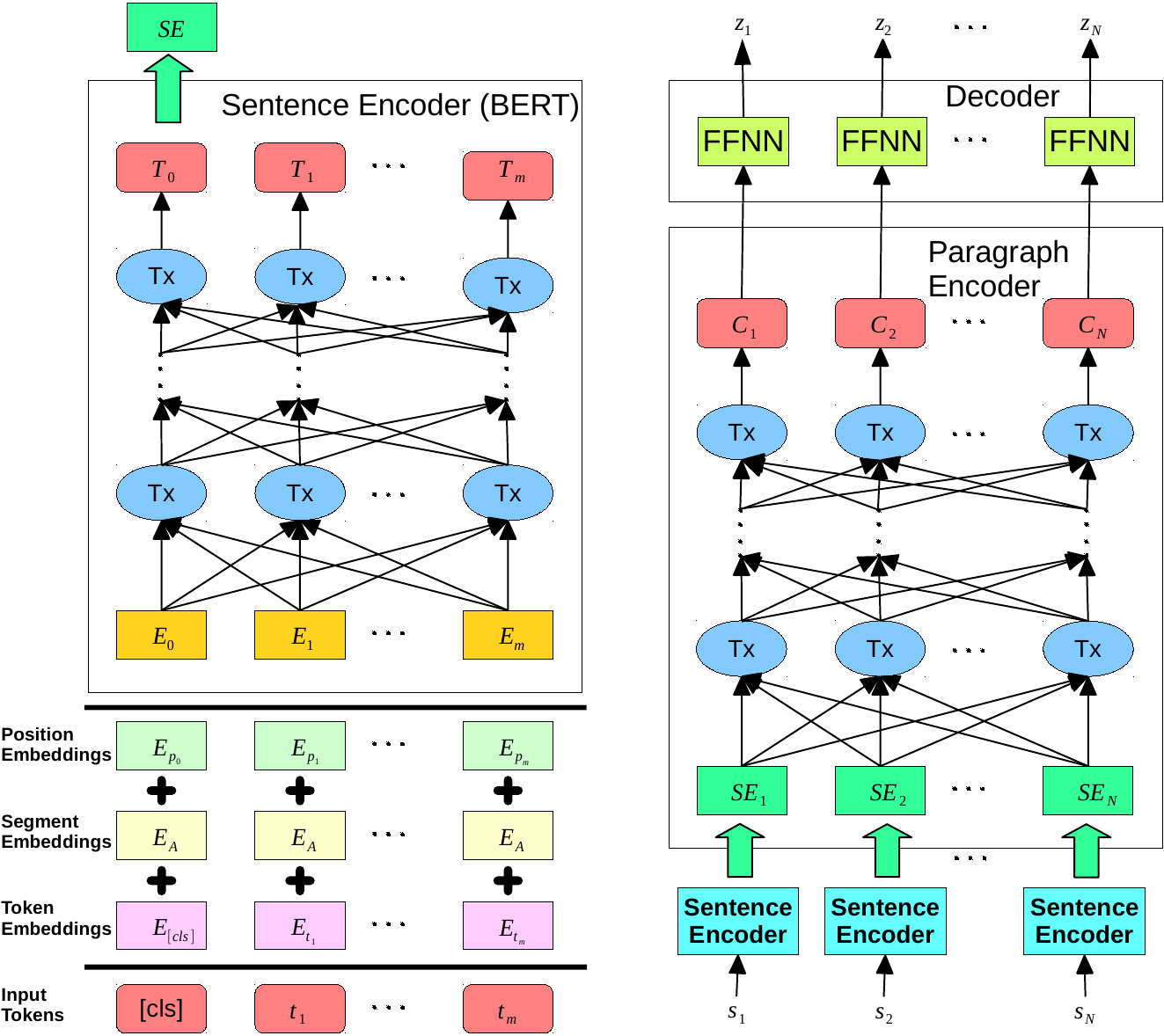}
        }\quad
    \subfloat{
    \includegraphics[width=0.3\textwidth]{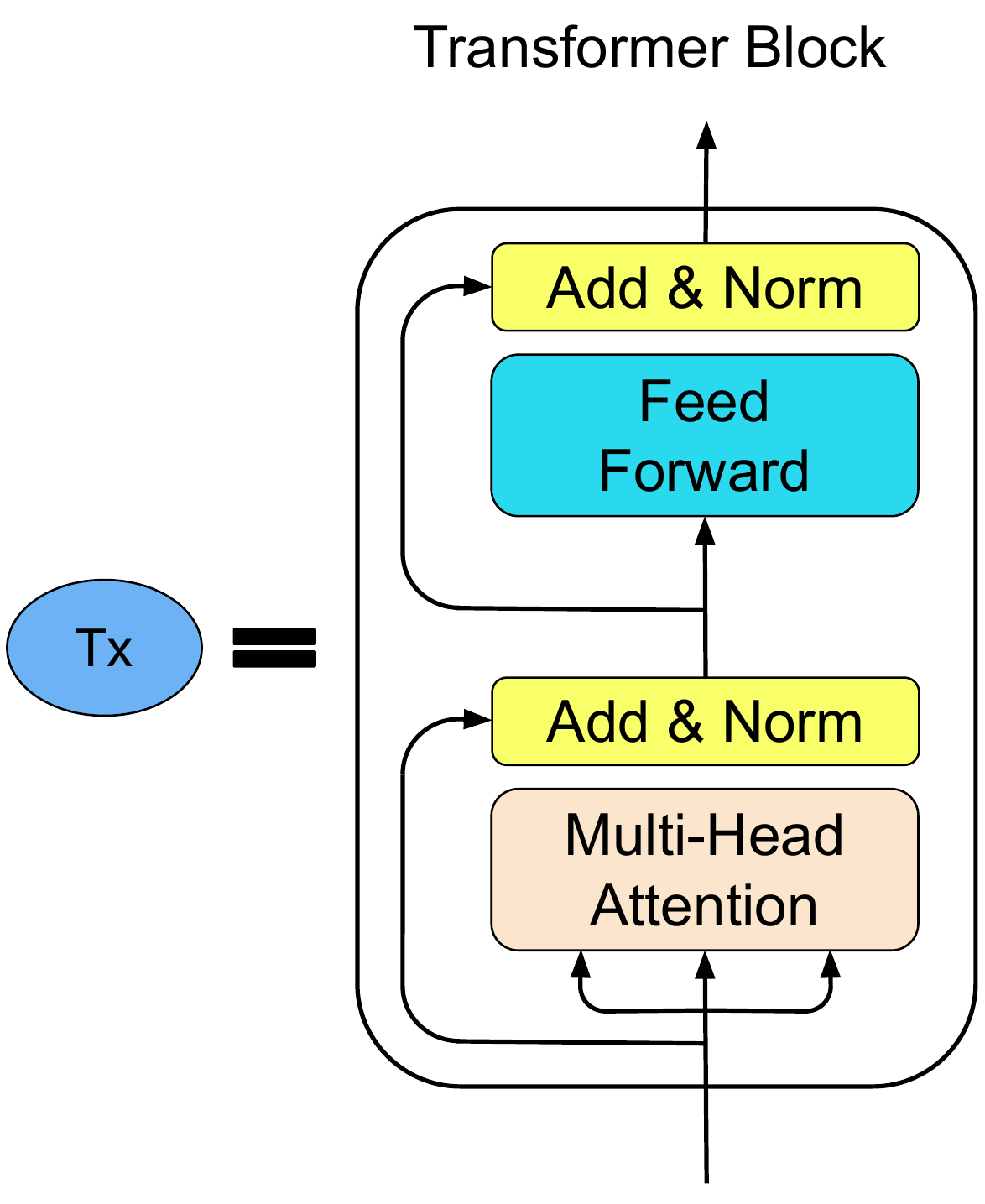}
        }
    \caption{The proposed model architecture}\label{fig:arch}
\end{figure*}

The proposed model has three components: a sentence encoder, a paragraph encoder and a decoder.
The sentence encoder is a Transformer \cite{vaswani2017attention} based pre-trained BERT \cite{bert}.
The paragraph encoder is a randomly initialized Transformer network.
The decoder is a simple feed forward neural network.  
A detailed diagram of the architecture is shown in Fig.~\ref{fig:arch}.

BERT as the sentence encoder allows us to use the language modeling knowledge obtained by pre-training on freely available, large, un-annotated text data.
The Transformer in the paragraph encoder ensures order invariant interaction between the input set of sentences, which is a random permutation of the desired paragraph.
The order invariant representation is a result of not adding positional encoding as well as the self-attention mechanism, which attends to every sentence encoding in the set with a direct connection. 
While in an LSTM encoder, sentences interact through recurrent connections, which limits the flow of information between sentences occurring farther in the sequence.

FFNN decoder provides a score corresponding to each sentence in the sentence set.
These scores are used for ranking the sentences to produce the correct ordering.
We train our model to compute the scores for all sentences, which gives us the flexibility to use the \emph{learning to rank} framework. Sorting these scores provides the correct ordering in the paragraph.
A variety of ranking loss functions available in the literature can be used for studying the trade-offs between different evaluation metrics.

We mention the details of the different components in our model below.

\subsection{Transformer Mechanism}
We briefly describe the transformer mechanism proposed by~\cite{vaswani2017attention}, used in the sentence encoder as well as the paragraph encoder.
Both of them use multiple self-attention layers.
Each layer has a multi-head self-attention sub-layer and a position wise feed-forward sub-layer.
These sub-layers use residual connections~\cite{he2016deep}, which allows easy passage of information through a deep stack of layers.
Layer normalization~\cite{lei2016layer}, $LayerNorm(x + Sublayer(x))$, is also used after each sub-layer, where $Sublayer(x)$ denotes the sub-layer function.

The attention mechanism is defined on queries, keys and values packed together in matrices $\textbf{Q}$, $\textbf{K}$ and $\textbf{V}$, respectively.
\begin{equation}
   \mathrm{Attention}(\textbf{Q}, \textbf{K}, \textbf{V}) = \mathrm{softmax}\left(\frac{\textbf{Q}\textbf{K}^\mathrm{T}}{\sqrt{\mathrm{d_k}}}\right)\textbf{V}
\end{equation}
A multi-head attention for query matrix $\textbf{Q}$, key matrix $\textbf{K}$ and value matrix $\textbf{V}$ is given by
\begin{equation}
    \mathrm{MultiHead}(\textbf{Q}, \textbf{K}, \textbf{V}) = \mathrm{Concat}(\textbf{H}_1, ..., \textbf{H}_h)\textbf{W}^O
\end{equation}
\begin{equation}
    \text{where }\textbf{H}_i = \mathrm{Attention}(\textbf{QW}^Q_i, \textbf{KW}^K_i, \textbf{VW}^V_i)
\end{equation}

Here, $\textbf{W}^Q_i \in \mathbb{R}^{d_{\mathrm{model}} \times d_k}$, $\textbf{W}^K_i \in \mathbb{R}^{d_{\mathrm{model}} \times d_k}$, $\textbf{W}^V_i \in \mathbb{R}^{d_{\mathrm{model}} \times d_v}$ and $\textbf{W}^O \in \mathbb{R}^{hd_v \times d_{\mathrm{model}}}$ are parameter matrices.
Every layer in the model outputs a vector of $d_{model}$ dimensions. $d_k$ and $d_v$ are dimensions of key and value, respectively, in a single head and there are $h$ such heads in total.
In self-attention, $\textbf{Q}$, $\textbf{K}$ and $\textbf{V}$ all are from the same layer.
In the sentence encoder each key, query and value is a vector corresponding to a word whereas in the paragraph encoder each such vector corresponds to a sentence.

\subsection{Sentence Encoder}
We use BERT for encoding each sentence in the paragraph. BERT is trained on an unsupervised Language Modeling (LM) task on English Wikipedia and BookCorpus~\cite{zhu2015aligning} datasets.
It is jointly optimized for two LM tasks, \emph{Masked LM} and \emph{Next Sentence Prediction}.
For \emph{Masked LM}, some random words in a sentence are replaced with either a mask word \emph{[MASK]} or a random word or kept unchanged. \emph{Masked LM} task aims to predict the masked word correctly.
\emph{Next Sentence Prediction} determines whether two sentences in the input appear next to each other.
In the process of learning these simple tasks on a large text corpus, BERT learns to represent sentences.
Pre-trained BERT can be further fine-tuned for other NLP tasks.
We use pre-trained BERT for sentence encoding and its parameters are also fine-tuned in an end-to-end manner for the  sentence ordering task.
BERT can be viewed as a multi-layer Transformer network with the layers as described in the sections above. 

\subsection{Paragraph Encoder}
To encode a paragraph, we again use a Transformer Network with self-attention layers. But the positional embedding layer in a standard Transformer is removed to handle the unordered nature of the sentences in the input paragraph.
The paragraph encoder outputs a $d_{model}$ dimensional vector for every sentence.
The encoded paragraph can be considered as a variable length vector of size $m_n \times d_{model}$, where $m_n$ is number of sentences in the $n^{th}$ paragraph.
Every node in the self-attention layer interacts with each node, including itself via length one connections. This ensures direct sharing of information among all the sentences in a paragraph.

\subsection{Decoder}
We use a position wise FFNN decoder that converts a sentence representation into a score, one score for each sentence.
Unlike in many previous works where decoding has to be done one sentence at a time, our model computes a score for each sentence in parallel.
Finally, the order of the paragraph is predicted by sorting these scores.

\subsection{Training and Ranking Losses}
Given a corpus with $N$ paragraphs, the $n^{th}$ paragraph with $m_n$ unordered sentences is denoted by $\mathbf{p}_n$ $=[s_1,s_2,\cdots,s_{m_n}]$. We denote the score of sentence $s_k$ by $z_k$, i.e., $z_k=f(s_k)$, where $f$ is the function denoting the model.
To train our model, we have used three types of loss functions.
The general form of the loss function is:
\begin{equation}
    Loss = \frac{1}{N}\sum_{n=1}^N \mathcal{L}(\mathbf{p}_n)
\end{equation}
where $\mathcal{L}(.)$ can be one of the following losses.

\textbf{Pointwise Ranking Loss:} In pointwise ranking approach, we view sentence ordering as a regression problem. Each sentence in an ordered paragraph is mapped onto a real-valued gold score $y_k \in [0,1]$, for sentence $s_k$ (increasing order). For example, if a paragraph has 5 sentences then $[y_1,y_2,y_3,y_4,y_5]=[0, 0.25, 0.50, 0.75, 1.0]$.
A similar approach for target values has been adopted by \cite{mcclure2018context}.
The training optimizes the MSE loss:
\begin{equation}
    \mathcal{L}_{\text{point}}(\mathbf{p}_n) = \frac{1}{m_n}\sum_{k=1}^{m_n} (y_k-z_k)^2
\end{equation}

\textbf{Pairwise Ranking Loss:} For pairwise ranking approach, we take the pairwise margin ranking loss~\cite{joachims2002optimizing} between two consecutive sentences while training. This does not require target value as in pointwise approach. We adopt the pairwise margin ranking loss in our approach as follows:
\begin{align}
    \mathcal{L}_{\text{pair}}(\mathbf{p}_n) &=\frac{1}{ m_n-1}\sum_{k=1}^{m_n-1}l_{p}(s_k,s_{k+1})\\
    l_{p}(s_k,s_{k+1}) &=\max(0,t.(z_k-z_{k+1})+\gamma)
\end{align}
where $t=1$ if sentence $s_k$ is placed in a higher position than sentence $s_{k+1}$ and $t=-1$ if sentence $s_{k+1}$ is in a higher position than $s_k$.
$\gamma$ is the margin hyperparameter.

\textbf{Listwise Ranking Loss:}
The listwise ranking loss considers all the sentences in a paragraph together. We have experimented with two different listwise losses: ListNet \cite{cao2007learning} and ListMLE \cite{xia2008listwise}. 

\begin{itemize}
    \item \textbf{ListNet}: This approach~\cite{cao2007learning} uses a loss based on the probability of a sentence being ranked on the top, given the scores of all the sentences in the paragraph. The top-one probability of sentence $s_k$ in the $n^{th}$ paragraph using gold target values and predicted scores are denoted by $P_n(s_k)$ and $\hat{P}_n(s_k)$, respectively, which are given as:
    \begin{equation}
        P_n(s_k) = \frac{\exp(y_k)}{\sum_{i=1}^{m_n} \exp(y_i)}
    \end{equation}
    \begin{equation}
    \hat{P}_n(s_k) = \frac{\exp(z_k)}{\sum_{i=1}^{m_n} \exp(z_i)}
    \end{equation}
    
    \begin{equation}
        \mathcal{L}_{\text{Net}}(\mathbf{p}_n)=
         -\sum_{k=1}^{m_n}P_n(s_k)\log \hat{P}_n(s_k)
    \end{equation}
    where $y_k \in [0,1]$ is the gold score for sentence $s_k$.
    \item \textbf{ListMLE}: In this method, the likelihood loss function is minimized, which is a surrogate loss to the perfect order based 0-1 loss function \cite{xia2008listwise}. Let the correct order of paragraph $n$ be $\mathbf{o}=[o_1,o_2,\cdots,o_{m_n}]$.
    \begin{align}
        \mathcal{L}_{\text{MLE}}(\mathbf{p}_n)=
        -\log P_M(\mathbf{o}|\mathbf{p}_n)\\
        P_M(\mathbf{o}|\mathbf{p}_n)=\prod_{k=1}^{m_n}
        \frac{\exp(z_{o_k})}{\sum_{i=k}^{m_n} \exp(z_{o_i})}
    \end{align}
    Since, ListMLE gives higher scores to sentences occurring in the starting positions (contrary to our other approaches), we reverse the order to make final predictions.
\end{itemize}

\section{Related work}
\textbf{Traditional Approaches:} \cite{Morris1991} use thesaurus for identifying lexical chains. \cite{Lapata2003} calculates transition probabilities between sentences and decode the sentence order greedily. \cite{barzilay2004catching} uses topic and topic transition modeling with Hidden Markov Model (HMM). \cite{barzilay2008modeling} extracts entities and  learns entity transition probabilities. A limitation of these approaches is their reliance on linguistic domain knowledge.
\\
\textbf{Deep learning approaches:} The recent trend has been towards using data-driven, end-to-end deep learning models for sentence ordering.~\cite{chen2016neural} predicts pairwise ordering (Pairwise Ranking Model) of sentences in the text and uses the predicted orderings in Window Network~\cite{li2014model}. It should be noticed that our method \ourmodel{} with pairwise loss differs from this model in the sense that we use entire context rather than pairs of sentences, independently.

Hierarchical deep learning models have dominated recent progress in the sentence ordering task.
These models have three key components: a Sentence Encoder, a Paragraph Encoder and a Decoder.
The sentence encoder processes words to get sentence vectors, then the paragraph encoder uses sentence vectors for computing a paragraph or context vector.
Further, these sentence and paragraph vectors are used by the decoder for computing the conditional probability of an ordering.
The network is trained for maximizing conditional probability of the correct ordering.

The hierarchical recurrent neural network (RNN) model of~\cite{gong2016end,logeswaran2018sentence} has both sentence encoder and  paragraph encoder based on LSTM.
The decoder is an LSTM based Pointer network, which, at every time step, predicts the next sentence probability for each candidate sentence.
At test time these probabilities are used in beam search for predicting the output order.
The LSTM paragraph encoder processes sentences in a given order, which makes it vulnerable to the order in which sentences are provided, which is a random permutation of the correct ordering.
So~\cite{cui-etal-2018-deep} replaced the LSTM based paragraph encoder with a Transformer~\cite{vaswani2017attention} (without positional encoding) based encoder, to make it order insensitive.
Recently, ~\cite{wang2019hierarchical} added a Transformer along with an LSTM in the sentence encoder to further improve the model. It is important to note that in this model, the Transformer in the sentence encoder cannot be initialized with a pre-trained BERT as it uses the output of LSTM layers rather than tokens from the input sentence. Most of the recent works use pre-trained word vectors for using linguistic knowledge from language models.

Ranking problems have been extensively studied in \emph{Learning to Rank} framework~\cite{burges2005learning} in many domains like information retrieval, machine translation, computational biology~\cite{Duh2008}, recommender systems~\cite{Lv2011} and software engineering~\cite{Xuan2014}.
Properties of ranking loss functions have been well studied.
They can be used for directly optimizing different properties of the sentence ordering task.
We use various ranking loss functions, which are capable of utilizing the relevance score of a single sentence for the global ranking of sentences, to optimize the network.

\begin{table*}[!htbp]
    \centering
    \resizebox{1.0\textwidth}{!}{
    \begin{tabular}{c|cc|cc|cc|cc|cc|cc} \toprule
        \multicolumn{1}{c}{\textbf{Methods}} & \multicolumn{2}{c}{\textbf{NIPS abstracts}} & \multicolumn{2}{c}{\textbf{AAN abstracts}} & \multicolumn{2}{c}{\textbf{NSF abstracts}} & \multicolumn{2}{c}{\textbf{arXiv abstracts}} & \multicolumn{2}{c}{\textbf{SIND captions}} & \multicolumn{2}{c}{\textbf{ROCStory}} \\ 
        \cmidrule(l{2pt}r{2pt}){2-3} \cmidrule(l{2pt}r{2pt}){4-5} \cmidrule(l{2pt}r{2pt}){6-7} \cmidrule(l{2pt}r{2pt}){8-9} \cmidrule(l{2pt}r{2pt}){10-11} \cmidrule(l{2pt}r{2pt}){12-13}
        {} & $\boldsymbol{\tau}$ & \textbf{PMR} & $\boldsymbol{\tau}$ & \textbf{PMR} &  $\boldsymbol{\tau}$ & \textbf{PMR}& $\boldsymbol{\tau}$ & \textbf{PMR} &  $\boldsymbol{\tau}$ & \textbf{PMR} &  $\boldsymbol{\tau}$ & \textbf{PMR}\\ 
        \hline
        {Entity Grid}  & {0.09}& {-}&  {0.10}& {-}&  {-}& {-}&  {-}& {-}& {-}&  {-}& {-}& {-} \\
        {Seq2seq}  & {0.27}& {-}&  {0.40}& {-}&  {0.10}& {-}&  {-}& {-}& {-}&  {-}& {-}& {-} \\
        {Window Network} & {0.59}& {-}&  {0.65}& {-}&  {0.28}&  {-}& {-}& {-}&  {-}& {-}& {-}& {-}\\
        {Pairwise Ranking Model} & {-}& {-}&  {-}& {-}&  {-}& {-}&  {0.66}& {33.43}&  {-}& {-}& {-}&  {-}\\
        {RNN Decoder} & {0.67}& {-}&  {0.66}& {-}&  {0.48}& {-}& {-}&  {-}& {-}& {-}& {-}&  {-}\\
        {Variant-LSTM+PtrNet} & {0.72}& {-}&  {0.73}& {-}&  {0.51}& {-}&  {-}& {-}& {-}&  {-}&  {-}& {-}\\
        {CNN+PtrNet} & {0.66}& {-}&  {0.69}& {-}&  {0.51}& {-}&  {0.71}& {39.28}&  {0.48}& {12.32}&  {-}& {-}\\
        {LSTM+PtrNet} & {0.67}& {-}&  {0.69}& {-}&  {0.52}& {-}& {0.72}& {40.44}&  {0.48}& {12.34}&  {0.7214}& {36.25}\\
        {ATTOrderNet} & {0.72}& {-}&  {0.73}& {-}&  {0.55}& {-}&  {0.73}& {42.19}&  {0.49}& {14.01}& {-}& {-}\\
        {HierarchicalATTNet} & {0.6671}& {14.06}&  {0.6903}& {31.29}& {0.5073}& {8.12}& {0.7536}& \textbf{44.55}& {0.5021}& {15.01}&  {0.7322}& \textbf{39.62}\\
        \hline
        \ourmodel{} {Regression} & {0.7324}& {18.91}&  {0.7472}& {35.61}&  {0.5607}& {9.03}&  {0.7449}& {39.85}&  {0.5510}& {14.03}& {0.7357}& {30.59}\\
        \ourmodel{} {Pairwise} & \textbf{0.7509}& {23.63}&  {0.7704}& {38.86}&  {0.5614}& {9.66}&  {0.7516}& {41.28}& {0.5609}& \textbf{15.59}& {0.7523}& {35.30}\\
        \ourmodel{} {ListNet} & {0.7463}& {23.88}&  {0.7644}& {37.51}&  {0.5772}& {9.48}& {0.7449}& {39.26}& {0.5507}& {14.18}& {0.7483}& {33.08}\\
        \ourmodel{} {ListMLE} & {0.7462}& \textbf{24.13}&  \textbf{0.7748}& \textbf{39.18}& \textbf{0.5798}& \textbf{9.78}& \textbf{0.7666}& {43.44}& \textbf{0.5652}& {15.48}& \textbf{0.7602}& {38.02}\\
        \bottomrule
    \end{tabular}
    }
    \caption{Kendall's tau ($\tau$) and perfect match ratio (PMR) on test set for various benchmark datasets. Note that Kendall's tau metric, on which our method consistently outperforms other baselines, correlates with human judgements.}
    \label{tab:sota}
\end{table*}

\begin{table}[!ht]
\centering
\resizebox{0.45\textwidth}{!}{
\begin{tabular}{lcccccccc} \toprule
\multicolumn{1}{c}{Datasets} & \multicolumn{1}{c}{Max} & \multicolumn{1}{c}{Avg} & \multicolumn{3}{c}{Dataset Split} \\
\cmidrule(l{2pt}r{2pt}){4-6} & & & Train & Val & Test \\
\cmidrule{1-6}
Accidents		& 19 & 11.5 & 100 & - & 100 \\ 
Earthquakes		& 32 & 10.4 & 100 & - & 99 \\ 
NIPS abstracts	& 15 & 6 & 2448 & 409 & 402 \\ 
AAN abstracts 	& 20 & 5 & 8569 & 962 & 2626 \\ 
NSF abstracts 	& 40 & 8.9 & 96070 & 10185 & 21580 \\
arXiv abstracts & 35 & 5.38 & 884912 & 110614 & 110615 \\
SIND captions   & 5 & 5 & 40155 & 4990 & 5055 \\
ROCStory        & 5 & 5 & 78529 & 9816 & 9817 \\
\bottomrule
\end{tabular}
}
\caption{Train, test and validation splits along with maximum and average paragraph lengths.}
\label{tab:dataset}
\end{table}

\section{Experiments}
We conduct a comprehensive analysis of our approach on various benchmark datasets and compare our model with other state-of-the-art approaches.
We also demonstrate the effectiveness of different components of our models by performing ablation analysis.

\subsection{Datasets}
Following \cite{cui-etal-2018-deep} and previous works we run our sentence ordering experiments on NIPS abstracts, AAN/ACL abstracts and NSF abstracts datasets from \cite{logeswaran2018sentence}; arXiv abstracts and SIND/VIST captions datasets from \cite{gong2016end,agrawal2016sort,huang2016visual}; and ROCStory dataset from \cite{wang2019hierarchical,mostafazadeh2016corpus}. Table \ref{tab:dataset} provides the statistics for each dataset. For order discrimination experiments, we use Accidents and Earthquakes datasets from \cite{barzilay2008modeling}.

\subsection{Hyperparameters}
For sentence encoder, we use the pre-trained BERT\textsubscript{BASE} model with 12 Transformer blocks, the hidden size as 768, and 12 self-attention heads. The feed-forward intermediate layer size is $4\times768$, i.e., $3072$.
The paragraph encoder is a Transformer Network having 2 Transformer blocks, with  hidden size 768 and a feed-forward intermediate layer size of $4\times768$, i.e., $3072$.
We experiment with 2, 4 and 8 Transformer blocks for ROCStory dataset; and 2 and 8 for arXiv dataset and report the best results.
The 768-dimensional sentence representation obtained from Transformer is pooled by the decoder which is a five layer feed-forward network with ReLU non-linearity in each layer with hidden size of 200, and a 1-dimensional output layer for the score. We train the model with Adam optimizer~\cite{kingma2014adam} with initial learning rate, $5\times10^{-5}$ for sentence encoder and paragraph encoder and $5\times10^{-3}$ for decoder; $\beta_1=0.9, \beta_2=0.999$; and batch size of $400$. For pairwise ranking loss, the value of the margin hyperparameter, $\gamma$, is set to 1.

\begin{figure*}[!tbp]
    \centering
    \includegraphics[width=0.95\linewidth]{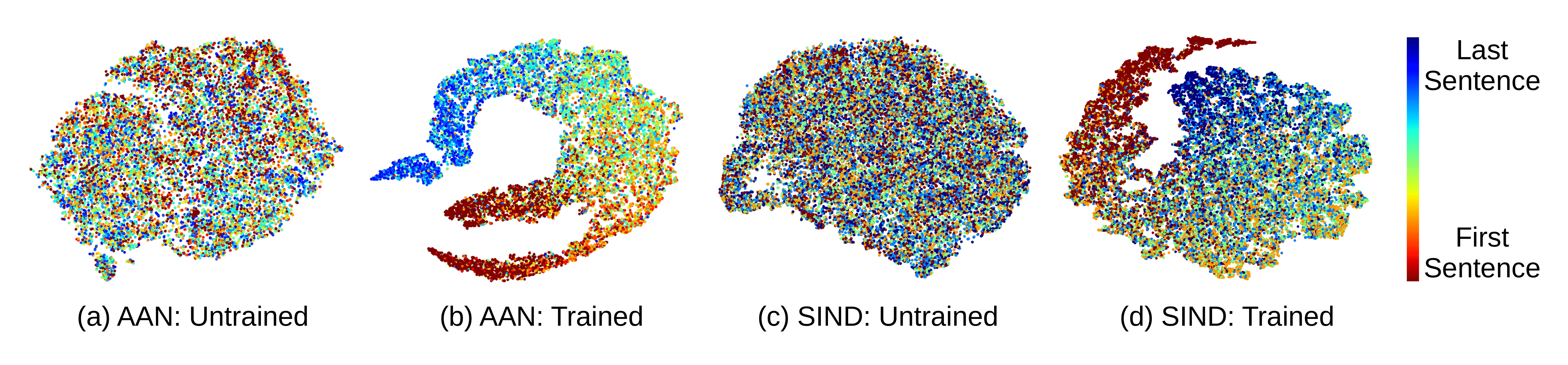}
    \caption{t-SNE embeddings of sentence representations for AAN abstracts and SIND captions datasets on sentence ordering task. Colors correspond to position of the sentences in the original paragraph. \textbf{Untrained:} Sentence embeddings before fine-tuning (same as pre-trained BERT). \textbf{Trained:} Sentence embeddings after fine-tuning. }\label{fig:tsne}
\end{figure*}

\subsection{Sentence Ordering}
\subsubsection{Evaluation}
Following previous studies we use the  following metrics for evaluating the sentence ordering task:

\textbf{Kendall's Tau ($\boldsymbol{\tau}$):}
For a sequence of length $n$, Kendall's tau ($\tau$) is defined as $\tau=1 - 2 \times (\#\ inversions) \mathbin{/} \binom nk$. \cite{lapata2006automatic} suggests that Kendall's tau score for sentence ordering correlates with human judgements.

\textbf{Perfect Match Ratio (PMR):} PMR is the fraction of number of exactly correct orderings over all the paragraphs. This is the toughest evaluation metric as it does not consider any partial match. Mathematically, it can be written as $\text{PMR}=\frac{1}{N}\sum_{n=1}^N \mathbb{I}\{\hat{o}^{(n)}=o^{*(n)}\}$, where $\hat{o}^{(n)}$ is the predicted order and $o^{*(n)}$ is the actual correct order for $n^{th}$ paragraph.

\subsubsection{Baselines}
We compare our methods with Entity Grid~\cite{barzilay2008modeling}, Seq2seq~\cite{li2014model}, Window Network~\cite{li2014model}, Pairwise Ranking Model~\cite{chen2016neural}, RNN Decoder, Variant-LSTM+PtrNet~\cite{logeswaran2018sentence}, CNN+PtrNet, LSTM+PtrNet~\cite{gong2016end}, ATTOrderNet~\cite{cui-etal-2018-deep} and HierarchicalATTNet~\cite{wang2019hierarchical}. Our methods are denoted by \ourmodel{} Regression, \ourmodel{} Pairwise, \ourmodel{} ListNet and \ourmodel{} ListMLE. 
\subsubsection{Results}
Table~\ref{tab:sota} provides the results of the sentence ordering experiments on six benchmark datasets. Most of the results of prior approaches have been taken from~\cite{cui-etal-2018-deep}. We consistently achieve better results than the state-of-the-art methods on $\tau$ on all the datasets. We reiterate that $\tau$ score correlates with human judgements. \ourmodel{} improves the previous state-of-the-art in terms of $\tau$ scores by the absolute percentage of $3.09\%$ in NIPS abstracts,  $4.48\%$ in AAN abstracts, $2.98\%$ in NSF abstracts, $1.3\%$ in arXiv abstracts, $6.31\%$ in SIND captions and $2.8\%$ in ROCStory dataset.
On SIND captions, we also get an improvement on PMR from $15.01\%$ to $15.59\%$. On all other datasets, we show competitive PMR score.
Among our approaches, pairwise and listwise methods always outperform the pointwise method. Specifically, ListMLE performs better than all other methods in most of the cases.

SIND contains descriptions of natural images and absence of visual information makes the ordering task more difficult compared to other datasets.
Higher maximum (40) and average (8.9) size of paragraphs make NSF a harder dataset than others, leading to lower performance.

More details about hyperparameters are provided in the supplementary material.

\begin{table}[h]
\centering
\resizebox{0.45\textwidth}{!}{
\begin{tabular}{lcccccccc} \toprule
\multicolumn{1}{c}{Methods} & \multicolumn{2}{c}{SIND} & \multicolumn{2}{c}{arXiv} \\
\cmidrule(l{2pt}r{2pt}){2-3} \cmidrule(l{2pt}r{2pt}){4-5} & First & Last & First & Last \\
\cmidrule{1-5}
Pairwise Ranking Model & - & - & 84.85 & 62.37 \\
CNN+PtrNet & 73.53 & 53.26 & 89.43 & 65.36 \\
LSTM+PtrNet & 74.66 & 53.30 & 90.47 & 66.49 \\
ATTOrderNet & 76.00 & 54.42 & 91.00 & 68.08 \\
\cmidrule{1-5}
\ourmodel{} Regression & 78.52 & 57.37 & 92.59 & 68.51 \\
\ourmodel{} Pairwise & 79.09 & 58.69 & 92.87 & \textbf{69.33} \\
\ourmodel{} ListMLE & \textbf{80.32} & \textbf{59.68} & \textbf{92.97} & 69.13 \\
\ourmodel{} Listnet & 78.00 & 58.18 & 92.46 & 68.64 \\
\bottomrule
\end{tabular}
}
\caption{Accuracy of predicting first and last sentences in SIND and arXiv datasets.}
\label{tab:firstlast}
\end{table}

As discussed by \cite{gong2016end,chen2016neural,cui-etal-2018-deep}, it is more significant to identify the first and the last sentences in a paragraph.
We provide the accuracies of our approach and compare it with the results reported in \cite{cui-etal-2018-deep} in Table~\ref{tab:firstlast} for SIND captions and arXiv abstracts datasets. 
All our approaches perform better than the state-of-the-art models.
In particular, on SIND dataset, our model achieves absolute improvement of $4.32\%$ and $5.26\%$ respectively, for first and last positions over the current state-of-the-art model.

\begin{figure*}[!ht]
    \centering
    \includegraphics[width=.75\textwidth]{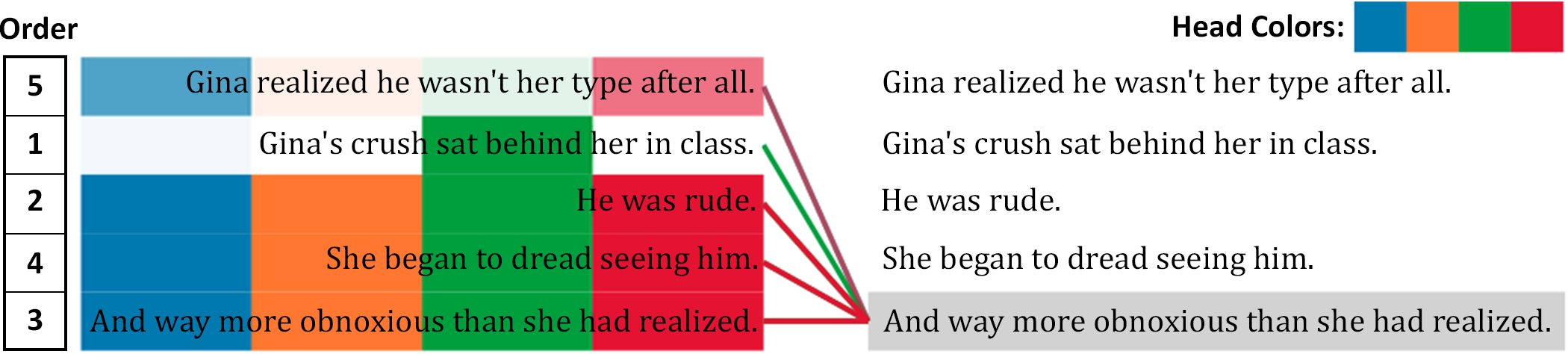}
     \caption{Self-attention scores of last layer of paragraph encoder: Attention score for all heads are shown in different colors. Higher Intensity represent higher values. First and last sentences can be differentiated, clearly. Sentences with positive and negative sentiments have different patterns in attention values.}\label{fig:sent}
\end{figure*}

To visualize the effect of training on the sentence representation, we use t-SNE embeddings. We show the t-SNE embeddings for AAN abstracts and SIND captions datasets before and after training in Fig.~\ref{fig:tsne}. It is important to note that before training, sentence representation is same as that of pre-trained BERT. The sentences are taken from the test set. Clearly, our model is learning to separate sentences in different positions and generalizes well on the unseen test set. This also shows that simply using the BERT representations is not enough as the t-SNE embeddings do not show any pattern before training the model on sentence ordering task. Figure~\ref{fig:sent} shows the visualization for paragraph encoder self-attentions. 

\begin{table}[!ht]
\centering
\resizebox{0.40\textwidth}{!}{
\begin{tabular}{lccc} \toprule
\multicolumn{1}{c}{Methods} & \multicolumn{1}{c}{Accidents} & \multicolumn{1}{c}{Earthquakes} \\
\cmidrule{1-3}
Graph & 84.6 & 63.5 \\
HMM+Entity & 84.2 & 91.1 \\
HMM & 82.2 & 93.8 \\
Entity Grid & 90.4 & 87.2 \\
Recurrent & 84.0 & 95.1 \\
Recursive & 86.4 & 97.6 \\
Discriminative Model & 93.0 & 99.2 \\
Variant-LSTM+PtrNet & 94.4 & 99.7 \\
LSTM+PtrNet & 93.7 & 99.5 \\
ATTOrderNet & 96.2 & \textbf{99.8} \\
\cmidrule{1-3}
\ourmodel{} & \textbf{96.79} & 98.65\\
\bottomrule
\end{tabular}
}
\caption{Performance of different models for order discrimination task on Accidents and Earthquakes datasets.}
\label{tab:order}
\end{table}

\subsection{Order Discrimination}
We present evaluation of our model on the Order Discrimination task defined in~\cite{barzilay2008modeling,elsner2008coreference,elsner2011extending}, in this section.
For a given paragraph and its randomly permuted sentences, the objective of the order discrimination task is to discriminate between the original and permuted paragraphs.
We analyze the models using percentage accuracy on this binary classification task.

For order discrimination we use the best sentence ordering model, found by validation on heldout data, to predict ordering for both original and permuted paragraphs.
We compute Kendall's Tau (higher if the number of inverted pairs is lower) for both predicted ordering, with respect to ordering $o = [0, 1, 2, \dots, m]$, where $m$ is the number of sentences in the paragraph.
The paragraph with higher Kendall's Tau value is classified as an original or more coherent paragraph.

Following~\cite{bertolino2005introducing,logeswaran2018sentence,cui-etal-2018-deep}, we evaluate our models on Accidents and Earthquakes datasets for order discrimination so that we can compare their performance with the current state-of-the-art models. We use 1986 and 1956 test pairs on Accidents and Earthquakes, respectively, with the same setup as in~\cite{barzilay2008modeling}. 

\subsubsection{Results}
We compare our models with  Graph~\cite{guinaudeau2013graph}, HMM and HMM+Entity~\cite{louis2012coherence}, Entity Grid~\cite{barzilay2008modeling}, Recurrent and Recursive~\cite{li2014model}, Discriminative model~\cite{li2016neural}, Variant-LSTM+PtrNet~\cite{logeswaran2018sentence}, CNN+PtrNet and
LSTM+PtrNet~\cite{gong2016end} and ATTOrderNet~\cite{cui-etal-2018-deep}.
Order discrimination results are reported in Table~\ref{tab:order}.

On Accidents, we outperform the current state-of-the-art and on Earthquakes, our model gives competitive results. We report our best performing method ListMLE on this task.

\subsection{Ablation Analysis}
\begin{figure}[!ht]
    \includegraphics[width=0.47\textwidth]{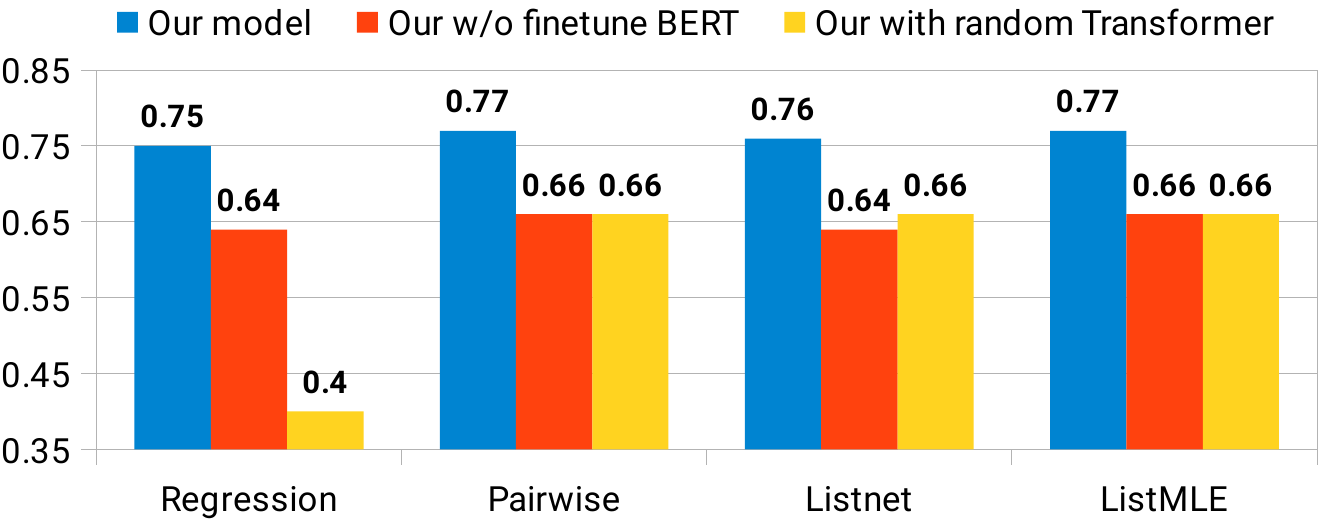}
    \begin{center}
        (a) AAN abstracts
    \end{center}
    \vfill
    \includegraphics[width=0.471\textwidth]{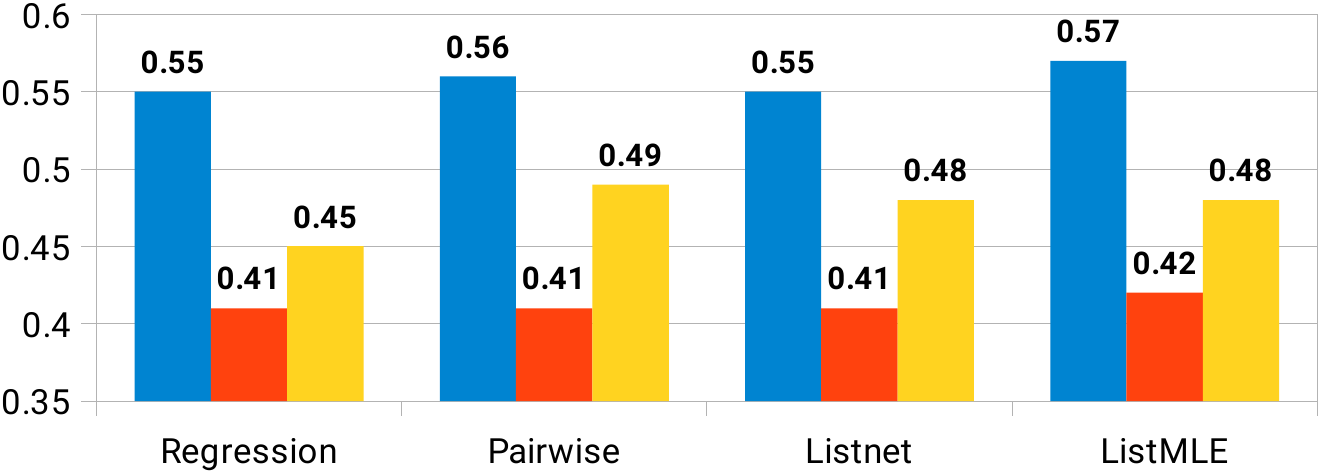}
    \begin{center}
        (b) SIND captions
    \end{center}
    \caption{Comparison of Kendall's tau ($\tau$) for our different models on AAN abstracts and SIND captions datasets with variants: i. Without fine-tuning BERT/sentence encoder, ii. Using randomly initialized BERT/sentence encoder.}
    \label{fig:abla}
\end{figure}

In this section, we conduct various ablation studies to assess our model and understand the roles played by different components in our model.
\\
\textbf{Finetune vs. No Finetune BERT}: To analyze how BERT is contributing as a sentence encoder without further training on the sentence ordering dataset, we make the parameters of BERT sentence encoder non-trainable. We conduct these experiments on AAN abstracts and SIND captions for all of our approaches. Fig.~\ref{fig:abla} shows that not fine-tuning BERT on the sentence ordering dataset reduces the performance drastically. This suggests that fine-tuning BERT/sentence encoder in an end-to-end fashion for this particular task contributes to significantly better learning.
\\
\textbf{Pretrained vs. Random Transformer}: To observe the effect of using BERT pre-trained model for sentence encoder, we run experiments on AAN abstracts and SIND captions using a randomly initialized Transformer and BERT initialized Transformer (our model). All weights in both the models are fine-tuned for the sentence ordering task. Fig.~\ref{fig:abla} shows that the performance reduces substantially for random Transformer. This indicates that our model is able to utilize the language modeling knowledge from pre-trained BERT for sentence ordering.

We observe similar ablations results for NIPS and ROC.
\section{Conclusion and Future work}
We propose a novel ranking loss and Transformer based \ourmodel{} for sentence ordering and order discrimination task.
\ourmodel{} uses pre-trained BERT for sentence representation and a Transformer for paragraph representation.
It uses a simpler feed forward network for decoding, compared to previous work.
The model can be trained on extensively studied ranking loss functions, opening a new direction for exploiting advances in the  ranking literature.
Our experiments demonstrate ability of the model to learn sentence ordering.
\ourmodel{} has improved the state-of-the-art results in sentence ordering and order discrimination tasks on various benchmark datasets.
We report Kendall's tau and perfect match ratio on six datasets for the sentence ordering task.
The simple feed forward network in the decoder can be easily replaced by any complex network. 
In a {\em learning to rank} setting different metrics have been optimized, directly.
Future work on sentence ordering can explore these methods.
Our model can be extended for other ordering problems.

\textbf{Acknowledgements:} The authors thank the anonymous reviewers for their feedback. PR acknowledges support from the Visvesvaraya Young Faculty Fellowship by MeitY, India. PK is supported by the Research-I Foundation at IIT Kanpur.

\bibliographystyle{aaai}
\bibliography{bibliography}

\begin{thebibliography}{}

\bibitem[\protect\citeauthoryear{Agrawal \bgroup et al\mbox.\egroup
  }{2016}]{agrawal2016sort}
Agrawal, H.; Chandrasekaran, A.; Batra, D.; Parikh, D.; and Bansal, M.
\newblock 2016.
\newblock Sort story: Sorting jumbled images and captions into stories.
\newblock {\em arXiv preprint arXiv:1606.07493}.

\bibitem[\protect\citeauthoryear{Barzilay and
  Elhadad}{2002}]{barzilay2002inferring}
Barzilay, R., and Elhadad, N.
\newblock 2002.
\newblock Inferring strategies for sentence ordering in multidocument news
  summarization.
\newblock {\em Journal of Artificial Intelligence Research}.

\bibitem[\protect\citeauthoryear{Barzilay and
  Lapata}{2008}]{barzilay2008modeling}
Barzilay, R., and Lapata, M.
\newblock 2008.
\newblock Modeling local coherence: An entity-based approach.
\newblock {\em Computational Linguistics}.

\bibitem[\protect\citeauthoryear{Barzilay and Lee}{2004}]{barzilay2004catching}
Barzilay, R., and Lee, L.
\newblock 2004.
\newblock Catching the drift: Probabilistic content models, with applications
  to generation and summarization.
\newblock In {\em HLT-NAACL}.
\newblock ACL.

\bibitem[\protect\citeauthoryear{Bertolino, Marchetti, and
  Muccini}{2005}]{bertolino2005introducing}
Bertolino, A.; Marchetti, E.; and Muccini, H.
\newblock 2005.
\newblock Introducing a reasonably complete and coherent approach for
  model-based testing.
\newblock {\em ENTCS}.

\bibitem[\protect\citeauthoryear{Burges \bgroup et al\mbox.\egroup
  }{2005}]{burges2005learning}
Burges, C.; Shaked, T.; Renshaw, E.; Lazier, A.; Deeds, M.; Hamilton, N.; and
  Hullender, G.~N.
\newblock 2005.
\newblock Learning to rank using gradient descent.
\newblock In {\em ICML}.

\bibitem[\protect\citeauthoryear{Cao \bgroup et al\mbox.\egroup
  }{2007}]{cao2007learning}
Cao, Z.; Qin, T.; Liu, T.-Y.; Tsai, M.-F.; and Li, H.
\newblock 2007.
\newblock Learning to rank: from pairwise approach to listwise approach.
\newblock In {\em ICML}.
\newblock ACM.

\bibitem[\protect\citeauthoryear{Chen, Qiu, and Huang}{2016}]{chen2016neural}
Chen, X.; Qiu, X.; and Huang, X.
\newblock 2016.
\newblock Neural sentence ordering.
\newblock {\em arXiv preprint arXiv:1607.06952}.

\bibitem[\protect\citeauthoryear{Cui \bgroup et al\mbox.\egroup
  }{2018}]{cui-etal-2018-deep}
Cui, B.; Li, Y.; Chen, M.; and Zhang, Z.
\newblock 2018.
\newblock Deep attentive sentence ordering network.
\newblock In {\em EMNLP}.
\newblock ACL.

\bibitem[\protect\citeauthoryear{Devlin \bgroup et al\mbox.\egroup
  }{2018}]{bert}
Devlin, J.; Chang, M.-W.; Lee, K.; and Toutanova, K.
\newblock 2018.
\newblock Bert: Pre-training of deep bidirectional transformers for language
  understanding.
\newblock {\em arXiv preprint arXiv:1810.04805}.

\bibitem[\protect\citeauthoryear{Duh and Kirchhoff}{2008}]{Duh2008}
Duh, K., and Kirchhoff, K.
\newblock 2008.
\newblock Learning to rank with partially-labeled data.
\newblock In {\em SIGIR}.
\newblock ACM.

\bibitem[\protect\citeauthoryear{Elsner and
  Charniak}{2008}]{elsner2008coreference}
Elsner, M., and Charniak, E.
\newblock 2008.
\newblock Coreference-inspired coherence modeling.
\newblock In {\em ACL-HLT}.
\newblock ACL.

\bibitem[\protect\citeauthoryear{Elsner and
  Charniak}{2011}]{elsner2011extending}
Elsner, M., and Charniak, E.
\newblock 2011.
\newblock Extending the entity grid with entity-specific features.
\newblock In {\em ACL-HLT}.

\bibitem[\protect\citeauthoryear{Gong \bgroup et al\mbox.\egroup
  }{2016}]{gong2016end}
Gong, J.; Chen, X.; Qiu, X.; and Huang, X.
\newblock 2016.
\newblock End-to-end neural sentence ordering using pointer network.
\newblock {\em arXiv preprint arXiv:1611.04953}.

\bibitem[\protect\citeauthoryear{Guinaudeau and
  Strube}{2013}]{guinaudeau2013graph}
Guinaudeau, C., and Strube, M.
\newblock 2013.
\newblock Graph-based local coherence modeling.
\newblock In {\em ACL}.

\bibitem[\protect\citeauthoryear{He \bgroup et al\mbox.\egroup
  }{2016}]{he2016deep}
He, K.; Zhang, X.; Ren, S.; and Sun, J.
\newblock 2016.
\newblock Deep residual learning for image recognition.
\newblock In {\em CVPR}.

\bibitem[\protect\citeauthoryear{Hochreiter and
  Schmidhuber}{1997}]{Hochreiter1997LSTM}
Hochreiter, S., and Schmidhuber, J.
\newblock 1997.
\newblock Long short-term memory.
\newblock {\em Neural Comput.}

\bibitem[\protect\citeauthoryear{Huang \bgroup et al\mbox.\egroup
  }{2016}]{huang2016visual}
Huang, T.-H.~K.; Ferraro, F.; Mostafazadeh, N.; Misra, I.; Agrawal, A.; Devlin,
  J.; Girshick, R.; He, X.; Kohli, P.; Batra, D.; et~al.
\newblock 2016.
\newblock Visual storytelling.
\newblock In {\em HLT-NAACL}.

\bibitem[\protect\citeauthoryear{Joachims}{2002}]{joachims2002optimizing}
Joachims, T.
\newblock 2002.
\newblock Optimizing search engines using clickthrough data.
\newblock In {\em ACM SIGKDD}.
\newblock ACM.

\bibitem[\protect\citeauthoryear{Kingma and Ba}{2014}]{kingma2014adam}
Kingma, D.~P., and Ba, J.
\newblock 2014.
\newblock Adam: A method for stochastic optimization.
\newblock {\em arXiv preprint arXiv:1412.6980}.

\bibitem[\protect\citeauthoryear{Konstas and Lapata}{2012}]{konstas2012concept}
Konstas, I., and Lapata, M.
\newblock 2012.
\newblock Concept-to-text generation via discriminative reranking.
\newblock In {\em ACL}.
\newblock ACL.

\bibitem[\protect\citeauthoryear{Lapata}{2003}]{Lapata2003}
Lapata, M.
\newblock 2003.
\newblock Probabilistic text structuring: Experiments with sentence ordering.
\newblock In {\em ACL}.
\newblock ACL.

\bibitem[\protect\citeauthoryear{Lapata}{2006}]{lapata2006automatic}
Lapata, M.
\newblock 2006.
\newblock Automatic evaluation of information ordering: Kendall's tau.
\newblock {\em Computational Linguistics}.

\bibitem[\protect\citeauthoryear{Lei~Ba, Kiros, and
  Hinton}{2016}]{lei2016layer}
Lei~Ba, J.; Kiros, J.~R.; and Hinton, G.~E.
\newblock 2016.
\newblock Layer normalization.
\newblock {\em arXiv preprint arXiv:1607.06450}.

\bibitem[\protect\citeauthoryear{Li and Hovy}{2014}]{li2014model}
Li, J., and Hovy, E.
\newblock 2014.
\newblock A model of coherence based on distributed sentence representation.
\newblock In {\em EMNLP}.

\bibitem[\protect\citeauthoryear{Li and Jurafsky}{2016}]{li2016neural}
Li, J., and Jurafsky, D.
\newblock 2016.
\newblock Neural net models for open-domain discourse coherence.
\newblock {\em arXiv preprint arXiv:1606.01545}.

\bibitem[\protect\citeauthoryear{Logeswaran and
  Lee}{2018}]{logeswaran2018efficient}
Logeswaran, L., and Lee, H.
\newblock 2018.
\newblock An efficient framework for learning sentence representations.
\newblock {\em arXiv preprint arXiv:1803.02893}.

\bibitem[\protect\citeauthoryear{Logeswaran, Lee, and
  Radev}{2018}]{logeswaran2018sentence}
Logeswaran, L.; Lee, H.; and Radev, D.
\newblock 2018.
\newblock Sentence ordering and coherence modeling using recurrent neural
  networks.
\newblock In {\em AAAI}.

\bibitem[\protect\citeauthoryear{Louis and Nenkova}{2012}]{louis2012coherence}
Louis, A., and Nenkova, A.
\newblock 2012.
\newblock A coherence model based on syntactic patterns.
\newblock In {\em EMNLP}.
\newblock ACL.

\bibitem[\protect\citeauthoryear{Lv \bgroup et al\mbox.\egroup }{2011}]{Lv2011}
Lv, Y.; Moon, T.; Kolari, P.; Zheng, Z.; Wang, X.; and Chang, Y.
\newblock 2011.
\newblock Learning to model relatedness for news recommendation.
\newblock In {\em WWW}.

\bibitem[\protect\citeauthoryear{McClure, O'Brien, and
  Roy}{2018}]{mcclure2018context}
McClure, D.; O'Brien, S.; and Roy, D.
\newblock 2018.
\newblock Context is key: New approaches to neural coherence modeling.
\newblock {\em arXiv preprint arXiv:1812.04722}.

\bibitem[\protect\citeauthoryear{Morris and Hirst}{1991}]{Morris1991}
Morris, J., and Hirst, G.
\newblock 1991.
\newblock Lexical cohesion computed by thesaural relations as an indicator of
  the structure of text.
\newblock {\em Comput. Linguist.}

\bibitem[\protect\citeauthoryear{Mostafazadeh \bgroup et al\mbox.\egroup
  }{2016}]{mostafazadeh2016corpus}
Mostafazadeh, N.; Chambers, N.; He, X.; Parikh, D.; Batra, D.; Vanderwende, L.;
  Kohli, P.; and Allen, J.
\newblock 2016.
\newblock A corpus and evaluation framework for deeper understanding of
  commonsense stories.
\newblock {\em arXiv preprint arXiv:1604.01696}.

\bibitem[\protect\citeauthoryear{Nallapati, Zhai, and
  Zhou}{2017}]{nallapati2017summarunner}
Nallapati, R.; Zhai, F.; and Zhou, B.
\newblock 2017.
\newblock Summarunner: A recurrent neural network based sequence model for
  extractive summarization of documents.
\newblock In {\em AAAI}.

\bibitem[\protect\citeauthoryear{Pennington, Socher, and
  Manning}{2014}]{pennington2014glove}
Pennington, J.; Socher, R.; and Manning, C.
\newblock 2014.
\newblock Glove: Global vectors for word representation.
\newblock In {\em EMNLP}.

\bibitem[\protect\citeauthoryear{Vaswani \bgroup et al\mbox.\egroup
  }{2017}]{vaswani2017attention}
Vaswani, A.; Shazeer, N.; Parmar, N.; Uszkoreit, J.; Jones, L.; Gomez, A.~N.;
  Kaiser, {\L}.; and Polosukhin, I.
\newblock 2017.
\newblock Attention is all you need.
\newblock In {\em NIPS}.

\bibitem[\protect\citeauthoryear{Verberne}{2011}]{verberne2011retrieval}
Verberne, S.
\newblock 2011.
\newblock Retrieval-based question answering for machine reading evaluation.
\newblock In {\em CLEF}.

\bibitem[\protect\citeauthoryear{Wang and Wan}{2019}]{wang2019hierarchical}
Wang, T., and Wan, X.
\newblock 2019.
\newblock Hierarchical attention networks for sentence ordering.
\newblock In {\em AAAI}.

\bibitem[\protect\citeauthoryear{Xia \bgroup et al\mbox.\egroup
  }{2008}]{xia2008listwise}
Xia, F.; Liu, T.-Y.; Wang, J.; Zhang, W.; and Li, H.
\newblock 2008.
\newblock Listwise approach to learning to rank: theory and algorithm.
\newblock In {\em ICML}.
\newblock ACM.

\bibitem[\protect\citeauthoryear{Xuan and Monperrus}{2014}]{Xuan2014}
Xuan, J., and Monperrus, M.
\newblock 2014.
\newblock Learning to combine multiple ranking metrics for fault localization.
\newblock In {\em ICSME}.

\bibitem[\protect\citeauthoryear{Yu \bgroup et al\mbox.\egroup
  }{2018}]{yu2018qanet}
Yu, A.~W.; Dohan, D.; Luong, M.-T.; Zhao, R.; Chen, K.; Norouzi, M.; and Le,
  Q.~V.
\newblock 2018.
\newblock Qanet: Combining local convolution with global self-attention for
  reading comprehension.
\newblock {\em arXiv preprint arXiv:1804.09541}.

\bibitem[\protect\citeauthoryear{Zhu \bgroup et al\mbox.\egroup
  }{2015}]{zhu2015aligning}
Zhu, Y.; Kiros, R.; Zemel, R.; Salakhutdinov, R.; Urtasun, R.; Torralba, A.;
  and Fidler, S.
\newblock 2015.
\newblock Aligning books and movies: Towards story-like visual explanations by
  watching movies and reading books.
\newblock In {\em CVPR}.

\end{thebibliography}

\end{document}